%% file: templateArxiv.tex
\documentclass{article}

\usepackage{PRIMEarxiv}

\usepackage{arabtex} 
\usepackage{utf8}
\setcode{utf8}
\usepackage{siunitx}  
\usepackage{hyperref}       
\usepackage{url}            
\usepackage{booktabs}       
\usepackage{amsfonts}       
\usepackage{subfig}         
\usepackage{nicefrac}       
\usepackage{float}          
\usepackage{microtype}      
\usepackage{lipsum}
\usepackage{natbib}
\usepackage{fancyhdr}       
\usepackage{graphicx}       
\graphicspath{{media/}}     

\title{Deciphering Emotions in Children Storybooks: A Comparative Analysis of Multimodal LLMs in Educational Applications}

\author{
  Bushra Asseri, Estabraq Abdelaziz, Maha Al Mogren, Tayef Alhefdhi, Areej Al-Wabil \\
  College of Engineering \& Advanced Computing \\
  Alfaisal University \\
  \texttt{\{basseri, eabaker, mmogren, talhefdhi01, awabil\}@alfaisal.edu}
}

\begin{document}
\maketitle

\begin{quote}
  \input{tex/abstract}  
\end{quote}


\section{Introduction}
Understanding emotions through visual forms is essential for effective communication and learning, especially in early childhood \citep{denham_early_2012,pons_emotion_2004}. As the role of Artificial Intelligence (AI) in education continues to grow, the ability to interpret emotions in illustrated children's books becomes increasingly important for developing culturally responsive educational technologies \citep{chen_application_2020}. Recent advances in MLLMs, such as GPT-4o and Gemini, have significantly improved visual-textual reasoning capabilities \citep{alayrac_flamingo_2022}, yet their application to non-English educational contexts remains limited \citep{joshi_state_2020}.
Arabic children's literature presents unique challenges for computational emotion recognition due to its expressive language, cultural symbolism, and artistic style \citep{habash_introduction_2010}. Despite the educational importance of emotion recognition in early literacy development highlighted by \cite{denham_plays_2010}, there exists a significant gap in our understanding of how well current AI systems can interpret emotional content in Arabic visual narratives.
This study addresses this research gap by systematically evaluating the emotion recognition capabilities of two advanced MLLMs, GPT-4o and Gemini Pro 1.5, when processing illustrations from Arabic children's storybooks. The primary aim of this study is to determine the accuracy and effectiveness of Large Language Models in identifying emotions in images from Arabic children's storybooks. To accomplish this aim, we pursue several specific objectives: to evaluate and compare the effectiveness of different prompting strategies (zero-shot, few-shot, chain-of-thought) on MLLMs' emotion recognition accuracy in Arabic visual narratives; to identify and categorize common error patterns when MLLMs interpret emotional content in Arabic children's literature; and to assess MLLMs' performance against human consensus when interpreting ambiguous or culturally nuanced emotional expressions.

To achieve these objectives, we compare the models' performance across three distinct prompting strategies using a dataset of 75 images from seven Arabic storybooks, with human-annotated emotions based on Plutchik's framework serving as ground truth \citep{plutchik_nature_2001}. Beyond simple accuracy metrics, we analyze cases where models struggle with narrative complexity or misinterpret subtle emotional expressions, providing insights into the challenges of cross-cultural emotion recognition in educational AI \citep{wei_chain--thought_2022}.
This research makes several contributions to the fields of educational technology and Arabic digital literacy: (1) it presents the first systematic evaluation of MLLMs for emotion recognition in Arabic children's literature; (2) it identifies specific patterns of success and failure in cross-cultural emotion recognition tasks; and (3) it provides actionable recommendations for developing more culturally sensitive AI systems for Arabic literacy education \citep{bender_data_2018}. Through exploratory experiments expanding the emotional label space and isolating individual characters from their narrative contexts, we offer additional insights into the models' emotional reasoning capabilities. The findings of this study have significant implications for the design of emotion-aware educational technologies that can effectively support Arabic literacy acquisition and social-emotional learning in culturally appropriate ways.

\section{Literature Review}

\subsection{Emotion Recognition in Educational Technologies}
Affective computing has facilitated the integration of emotion recognition within educational technology, enabling adaptive systems to personalize interactions by detecting engagement, confusion, frustration, and joy. Research shows that intelligent tutoring systems with emotion recognition capabilities increase student engagement and retention through responsive feedback. Recent studies highlight multimodal approaches that combine facial expressions, eye movements, and biosignal data, with empirical evidence showing significant improvements in learning outcomes when integrated into educational environments \citep{al-laith_monitoring_2021}.
These technological applications are grounded in established educational frameworks. Vygotsky's concept of the Zone of Proximal Development aligns with how adaptive systems adjust learning tasks based on detected emotional states \citep{vygotsky_mind_1978}, while Piaget's theory of cognitive development intersects with emotional development through the influence of emotional states on attention and focus \citep{piaget_intelligence_1981,denham_plays_2010}. These theoretical connections are especially relevant in cross-cultural contexts such as Arabic educational settings, where sociocultural factors significantly influence emotional expression and interpretation \citep{habash_introduction_2010, matsumoto_evidence_2011}.
\subsection{Arabic Children's Literature and Emotional Content}
Arabic children's literature presents unique challenges for computational emotion recognition due to its expressive language, cultural symbolism, and implicit emotional cues. The poetic tradition in Arabic literature contributes to its emotional complexity through specific formal structures and elaborate metaphors, with distinctive symbolism: desert imagery representing resilience, nature elements mirroring emotional states, and love metaphors expressing profound longing. This rich symbolic language creates challenges for computational systems trained primarily on literal expressions.
Additional challenges include the morphological complexity and diglossic nature of Arabic, where Modern Standard Arabic coexists with more than 25 regional dialects that differ morphologically, phonologically, and lexically. Children's literature often contains elements of both formal and dialectal forms, particularly in dialogue, creating a complex linguistic landscape for emotion detection \citep{alhuzali_spanemo_2021}.
Visual elements often convey emotional subtleties within Arabic children's stories, highlighting the necessity for AI models capable of nuanced multimodal understanding. Recent computational approaches have developed specialized Arabic emotion lexicons for basic emotions, demonstrating the need for culturally adapted resources \citep{abdul-mageed_emonet_2017}. Preliminary studies suggest that existing LLMs sometimes misinterpret culturally contextual imagery, underscoring the importance of culturally anchored datasets and annotations \citep{abdul-mageed_arbert_2021,alhuzali_spanemo_2021}.
\subsection{Plutchik's Emotion Model in Cross-Cultural Contexts}
Plutchik's Wheel of Emotions, comprising eight fundamental emotions (joy, trust, fear, surprise, sadness, disgust, anger, anticipation), serves as our analytical framework \citep{plutchik_nature_2001,mohammad_crowdsourcing_2013}. Each emotion is represented at three intensity levels, with opposites arranged diametrically and potential combinations identified (e.g., joy + trust = love) \citep{plutchik_nature_2001}.
However, the model's presumed universality requires scrutiny given cultural variability in emotional expression. For instance, emotions like "trust" ({\scriptsize\<ثقة>}) within Arabic narratives encompass broader concepts of reassurance, confidence, and relational sensibilities, diverging from Plutchik's Western-centric definitions \citep{al-laith_monitoring_2021}. When applied to Arabic emotion detection, researchers have adapted the framework by developing specialized seed lexicons corresponding to the eight primary emotions \citep{al-laith_monitoring_2021,nfaoui_evaluating_2024}.

Cross-cultural studies provide further evidence of both universality and cultural specificity in emotion recognition. \cite{chronaki_development_2018} demonstrated that while basic emotions can be recognized across unfamiliar languages, Arabic stimuli consistently yielded the lowest recognition rates among English listeners, with 92\% of participants rating Arabic as the most challenging language for emotion recognition. This indicates that while core emotional expressions may be universally recognizable, cultural and linguistic factors significantly influence their interpretation.
\subsection{Multimodal LLMs and Prompting Techniques}
State-of-the-art multimodal LLMs such as GPT-4o and Gemini demonstrate varying proficiencies in emotional interpretation within complex, narrative-rich scenarios. GPT-4o processes audio, vision, and text inputs in real-time, while Gemini models handle similar multimodal inputs with their "Deep Think" reasoning mode enhancing nuanced analysis \citep{alayrac_flamingo_2022}.
Recent evaluations reveal that Gemini achieves high accuracy in distinguishing emotional polarities across standard datasets but struggles with neutral expressions and complex academic emotions \citep{kasneci_chatgpt_2023}. Similarly, GPT models face challenges with Arabic morphology and syntax, highlighting the need for language-specific adaptation \citep{nfaoui_evaluating_2024}.
Performance significantly depends on the prompting methodology employed. Studies suggest that chain-of-thought (CoT) prompting, which instructs models to "think step-by-step," can enhance precision but may occasionally induce overinterpretation \citep{wei_chain--thought_2022}. Interestingly, research on Arabic emotion classification has found that prompts in English sometimes outperform Arabic prompts when addressing Arabic content, highlighting the complex interplay of language, culture, and model training \citep{nfaoui_evaluating_2024}.
These models employ various architectures for multimodal processing, from early-fusion approaches that interleave image and text tokens from initial layers to dual-encoder designs that separately process different modalities before integration \citep{alayrac_flamingo_2022}. The continued advancement of large context windows enables analysis of extended narratives and multi-page illustrations, critical for processing children's literature \citep{alayrac_flamingo_2022,team_gemini_2023}.
\subsection{Educational Applications and Arabic Literacy Development}
Integration of emotion-aware AI technologies in Arabic educational platforms remains limited despite their potential to enhance reading applications by dynamically customizing content delivery based on emotional feedback. Empirical studies demonstrate promising outcomes when emotion recognition is embedded in educational systems. A 2023 study analyzing schoolchildren's emotions and handwriting performance of Arabic letters showed that biofeedback on emotional states significantly improved learning outcomes \citep{zakraoui_study_2023}.
Arabic language textbooks are increasingly incorporating social-emotional learning components through texts, illustrations, and activities that promote self-understanding and respectful dialogue \citep{durlak_impact_2011}. However, these static materials lack the adaptive capabilities that AI-powered emotion recognition could provide. Research indicates that integrating emotional intelligence approaches in Arabic language learning yields substantial benefits for students' overall development.
The current scarcity of annotated Arabic datasets and the underrepresentation of Arabic-specific content in pretraining corpora underline the importance of targeted dataset development \citep{al-laith_monitoring_2021,nfaoui_evaluating_2024}. Recent initiatives like ArPanEmo address this gap by providing manually labeled Arabic content for multiple emotion categories \citep{althobaiti_open-source_2023}. Such resources are crucial for developing culturally appropriate emotion recognition systems for educational contexts in Arabic-speaking populations.
\subsection{Cultural Considerations and Design Recommendations}
Cultural context is critical in interpreting emotional expressions within educational AI applications. Systems lacking cultural sensitivity risk misinterpreting authors' intentions and fostering misunderstandings \citep{bender_data_2018}. Research in Arabic children's literature in educational settings indicates that original Arabic works often emphasize interpersonal emotional skills related to immediate social bonds rather than individual emotional competencies \citep{durlak_impact_2011}. This cultural orientation should be reflected in educational AI systems designed for Arabic-speaking children.
Designing culturally responsive educational AI requires prioritizing explainability, multimodal processing capabilities, and native language interactions. Recent advances in Arabic-specific natural language processing provide promising foundations, with models such as AraBERT, MARBERT, and QARiB demonstrating significant improvements in emotion classification when fine-tuned on task-specific data \citep{al-laith_monitoring_2021,nfaoui_evaluating_2024}.
Multimodal approaches show particular promise, combining text analysis with visual processing to capture the rich emotional content conveyed through illustrations in children's books. By integrating these modalities, systems can better understand the complementary or occasionally contradictory emotional signals present in text and images \citep{alhuzali_spanemo_2021}.
\subsection{Research Gaps and Study Rationale}
This review of the literature emphasizes the critical intersection of AI technologies, emotion recognition, and culturally nuanced Arabic educational contexts. Recent advances in multimodal LLMs, Arabic-specific transformer models, and emotion detection methodologies offer promising foundations for the development of culturally appropriate systems. However, significant gaps remain in understanding how these technologies can effectively interpret the complex emotional content of Arabic children's literature.
The development of Arabic-specific resources has accelerated in recent years, with new datasets like ArPanEmo (11,128 posts, ten emotions) \citep{althobaiti_open-source_2023}, LAMA (8,000 tweets, eight Plutchik emotions) \citep{al-laith_monitoring_2021}, and emotion lexicons providing valuable resources for training and evaluation. Similarly, transformer models pretrained on massive Arabic corpora, including MARBERT (128 GB, 50\% tweets) and QARiB (420M tweets + additional text), demonstrate significant improvements in Arabic language understanding. However, these resources primarily focus on social media content rather than children's literature, leaving a critical gap in age-appropriate and educationally relevant emotional content.
Cross-lingual approaches offer another promising direction, with recent research showing that translating English training data into Arabic and fine-tuning Arabic-specific models can achieve up to 90.9\% relative effectiveness compared to models trained directly on Arabic data. Such approaches could help address resource scarcity while maintaining cultural nuance.
To fully realize the potential of emotion-aware technologies within Arabic instructional settings, advancement in annotation frameworks, culturally responsive prompting techniques, and educational application design is essential. Future research should focus on:
\begin{itemize}
    \item Developing specialized datasets of Arabic children's literature annotated for emotional content
    \item Creating culturally calibrated emotion models that account for Arabic-specific expressions
    \item Designing and evaluating educational applications that leverage emotional understanding to enhance Arabic literacy development
    \item Establishing robust evaluation methodologies that incorporate cultural expertise and educational outcomes
\end{itemize}

Integrating emotionally intelligent AI demands shifting emphasis from technical excellence toward accommodating cultural and educational nuances, thereby enriching learners' emotional experiences. Given these insights and identified research gaps from existing literature, our methodological approach is explicitly designed to systematically address these challenges and enhance emotion recognition within Arabic children's literature.

\section{Materials and Methods}

This study employed a multi-method comparative analysis to evaluate the emotion recognition capabilities of Multimodal Large Language Models (MLLMs) when interpreting illustrations from Arabic children's literature. The methodological framework systematically integrated quantitative performance metrics with qualitative analytical techniques, facilitating rigorous comparison between human annotator judgments and machine-generated interpretations across diverse emotional contexts, prompting strategies, and image characteristics.

\subsection{Dataset}
The visual stimuli were sourced from "We Love Reading" ({\scriptsize\<نحن نحب القراءة>}) , an organization advancing Arabic literacy through culturally-relevant children's literature \citep{taghyeer_association_we_2023}. Seven distinct storybooks were randomly selected, with a minimum of ten illustrations systematically extracted from each, yielding 75 unique image panels. The images were distributed proportionally based on the narrative complexity and emotional diversity within each book, ensuring representative coverage across the collection. Selection criteria prioritized diverse emotional representations, varying degrees of emotional complexity, contextual clarity, and cultural specificity. Images were preserved in their entirety without segmentation to maintain ecological validity in accordance with established principles for multimodal research \citep{russell_circumplex_1980}.

All illustrations underwent standardization to ensure compatibility with MLLM vision processing requirements while preserving original visual information, consistent with methodological recommendations for ecologically valid evaluations of multimodal AI systems.

\subsection{Human Annotation}
Four annotators who are native Arabic speakers with full proficiency in Modern Standard Arabic and colloquial variants established ground truth classifications. Annotators received standardized instructions regarding the emotion taxonomy and classification protocols, following established guidelines for cultural annotation tasks \citep{hovy_social_2016}.

The emotion classification framework employed Plutchik's Wheel of Emotions taxonomy \citep{plutchik_nature_2001}, and was structured around nine distinct affective Arabic categories: happiness ({\scriptsize\<سعادة>}), sadness ({\scriptsize\<حزن>}), anger ({\scriptsize\<غضب>}), fear ({\scriptsize\<خوف>}), surprise ({\scriptsize\<مفاجأة>}), disgust ({\scriptsize\<قرف>}), neutral ({\scriptsize\<محايد>}), anticipation ({\scriptsize\<ترقب>}), and trust ({\scriptsize\<ثقة>}). This taxonomy was selected for its cultural adaptability within Arabic children's literature contexts \citep{abdul-mageed_emonet_2017}.

In instances of classificatory divergence, a structured consensus-building procedure was implemented where annotators met to discuss their interpretations. During these consensus meetings, annotators articulated interpretive rationales, examined visual evidence, and reconciled discrepant classifications through collaborative dialogue. This approach ensured ground truth represented informed intersubjective agreement rather than statistical aggregation and fostered deeper consideration of culturally nuanced emotional expressions.

\subsection{Selection and Interaction with MLLMs}
Two state-of-the-art MLLMs, OpenAI's GPT-4o and Google's Gemini 1.5 Pro, were selected based on their demonstrated performance on contemporary multimodal benchmarks and recognition for advanced multimodal understanding capabilities \citep{li_blip_2022}. Interactions with both models were conducted programmatically through their respective APIs (OpenAI API v1 and Google Gemini API), with special attention to maintaining consistent prompt formatting and presentation across all experimental conditions. We developed a custom Python framework to automate interactions with both APIs, ensuring methodological consistency and enabling systematic data collection. The API-based approach allowed for precise control over model parameters, systematic response collection, and reproducibility of results. All interactions occurred exclusively in Modern Standard Arabic, consistent with methodological standards for cross-lingual evaluation.

\subsection{Prompting Techniques}
Three distinct prompting paradigms were systematically implemented. In all prompting conditions, models were explicitly instructed in Modern Standard Arabic to select their response from a predefined list comprising Plutchik's eight primary emotions plus neutral to capture images lacking distinct emotional content:

\begin{enumerate}
    \item Zero-Shot Prompting involved direct instructions to identify the primary emotion from the predefined list without exemplification or methodological guidance. This approach tested models' baseline capabilities without contextual support.
    \item Few-Shot Prompting provided three image-emotion pairs exemplifying diverse emotional categories from the predefined list, following few-shot learning principles in contemporary vision-language research. This method examined whether exemplars enhanced recognition accuracy.
    \item Chain-of-Thought (CoT) Prompting incorporated explicit direction to "think step-by-step," engaging in sequential reasoning to identify visual cues, integrate observations, and classify emotions from the predefined list \citep{wei_chain--thought_2022}. This approach evaluated whether structured reasoning improved performance on emotionally complex stimuli.
\end{enumerate}

\subsection{Data Collection and Processing}
Responses from both GPT-4o and Gemini 1.5 across all prompting strategies were systematically collected for each image, yielding 450 machine-generated classifications (75 images × 2 models × 3 prompting strategies = 450 total classifications). Our automated data collection pipeline captured and stored model responses in a structured database with verification procedures to ensure accuracy. Each response was programmatically validated for conformity to the expected response format and manually reviewed when necessary.

Model outputs were standardized by aligning variant terms strictly to the predefined taxonomy through explicit mapping procedures. This standardization process included normalizing Arabic text variations, removing diacritics, and resolving synonym usage to ensure consistent emotional categorization. For CoT responses, only the final classifications were extracted for analytical comparison, adhering to standardized evaluation protocols for multimodal emotion recognition \citep{ghosal_dialoguegcn_2019}.

\subsection{Analysis Methods}
We employed both quantitative and qualitative analytical techniques to evaluate model performance:

\begin{enumerate}
    \item Performance Metrics: Overall performance, per-emotion precision, recall, and F1 scores were calculated for each model and prompting strategy by comparing model predictions against human-annotated ground truth.
    \item Error Analysis Framework: Errors were systematically categorized into three taxonomic categories: valence inversions (confusing positive/negative emotions), arousal confusions (misclassifying activation levels), and contextual/cultural misinterpretations, following established emotion recognition evaluation frameworks \citep{mohammad_understanding_2018}.
    \item Qualitative Case Studies: Representative examples of successful and unsuccessful classifications were subjected to in-depth qualitative analysis to identify patterns in model reasoning and cultural interpretation.
    \item Human-AI Alignment Analysis: Agreement between model predictions and human annotations was assessed using Cohen's Kappa to measure inter-rater reliability, accounting for chance agreement and class imbalance in emotion annotation tasks.
\end{enumerate}

\subsection{Supplementary Methodological Variations}
To further investigate factors influencing emotion recognition performance, we conducted two supplementary analyses using a representative subset (10\%) of the original dataset. Images for this analysis were selected to maintain proportional representation of emotional categories and visual complexity levels. For the first variation, we augmented standard prompting by including an image of Plutchik's wheel of emotions directly within the prompt interface, providing models with a visual reference framework of the emotional taxonomy. This approach aimed to assess whether explicit visualization of emotional relationships would improve classification performance.

In the second variation, we implemented character-focused segmentation, isolating only the main characters in each illustration while removing contextual backgrounds and surrounding elements. This method examined whether focusing models' attention on facial expressions and body language, without potentially distracting environmental cues, would enhance recognition performance. Both variations maintained identical prompting strategies (zero-shot, few-shot, and chain-of-thought) and evaluation procedures to enable direct comparison with our primary methodology. All supplementary analyses were conducted in Modern Standard Arabic with the same predefined list of nine emotion categories.

\section{Results}

The experimental analysis yielded a comprehensive dataset of (N=450) distinct emotion predictions (N = 75 images × 2 models × 3 prompting techniques) derived from the evaluation of multimodal large language models in Arabic emotional content recognition. This substantial corpus of predictions facilitated robust statistical assessment across varied experimental conditions. The investigation examined the differential performance of two state-of-the-art multimodal architectures (GPT-4o and Gemini) utilizing three distinct prompting paradigms: zero-shot, few-shot, and chain-of-thought methodologies. The dataset comprised storybook illustrations annotated across nine discrete emotional categories by native Arabic speakers, with a notable distributional asymmetry, as delineated in Table~\ref{tab:distribution}, wherein happiness ({\scriptsize\<سعادة>}) constituted 40\% of the annotations. This class imbalance is essential to consider when interpreting model performance. The analytical framework systematically addressed four principal dimensions: comparative efficacy of prompting strategies, emotion-specific classification performance, structured analysis of misclassification patterns through valence-arousal theoretical constructs, and contextual performance variations related to narrative positioning and visual ambiguity. The following sections present the quantitative and qualitative findings extracted from this substantial corpus of model predictions.

\begin{center}
\captionof{table}{Distribution of emotion labels in the Storybook dataset}
\label{tab:distribution}
\small   
\setlength{\tabcolsep}{6pt}        
\renewcommand{\arraystretch}{0.9}  
\begin{tabular*}{\linewidth}{@{\extracolsep{\fill}}%
  l                                    
  S[table-format=2.0]                  
  S[table-format=2.2]}                 
\toprule
\textbf{Annotation Emotion} & \textbf{Count} & \textbf{Percentage (\%)}\\
\midrule
Happiness ({\scriptsize\<سعادة>})      & 30 & 40.00\\
Anticipation ({\scriptsize\<ترقب>})   &  9 & 12.00\\
Neutral ({\scriptsize\<محايد>})       &  8 & 10.70\\
Trust ({\scriptsize\<ثقة>})           &  8 & 10.70\\
Sadness ({\scriptsize\<حزن>})         &  6 &  8.00\\
Fear ({\scriptsize\<خوف>})            &  6 &  8.00\\
Surprise ({\scriptsize\<مفاجأة>})     &  5 &  6.70\\
Anger ({\scriptsize\<غضب>})           &  2 &  2.70\\
Disgust ({\scriptsize\<قرف>})         &  1 &  1.30\\
\bottomrule
\end{tabular*}
\end{center}

\subsection{Prompting Technique Effectiveness}

To assess their performance in Arabic emotion recognition, results are presented in terms of overall performance and per-emotion classification performance.
\subsubsection{Overall Performance}

\begin{figure}[t]
    \centering
    \scriptsize
    \includegraphics[width=0.60\textwidth]{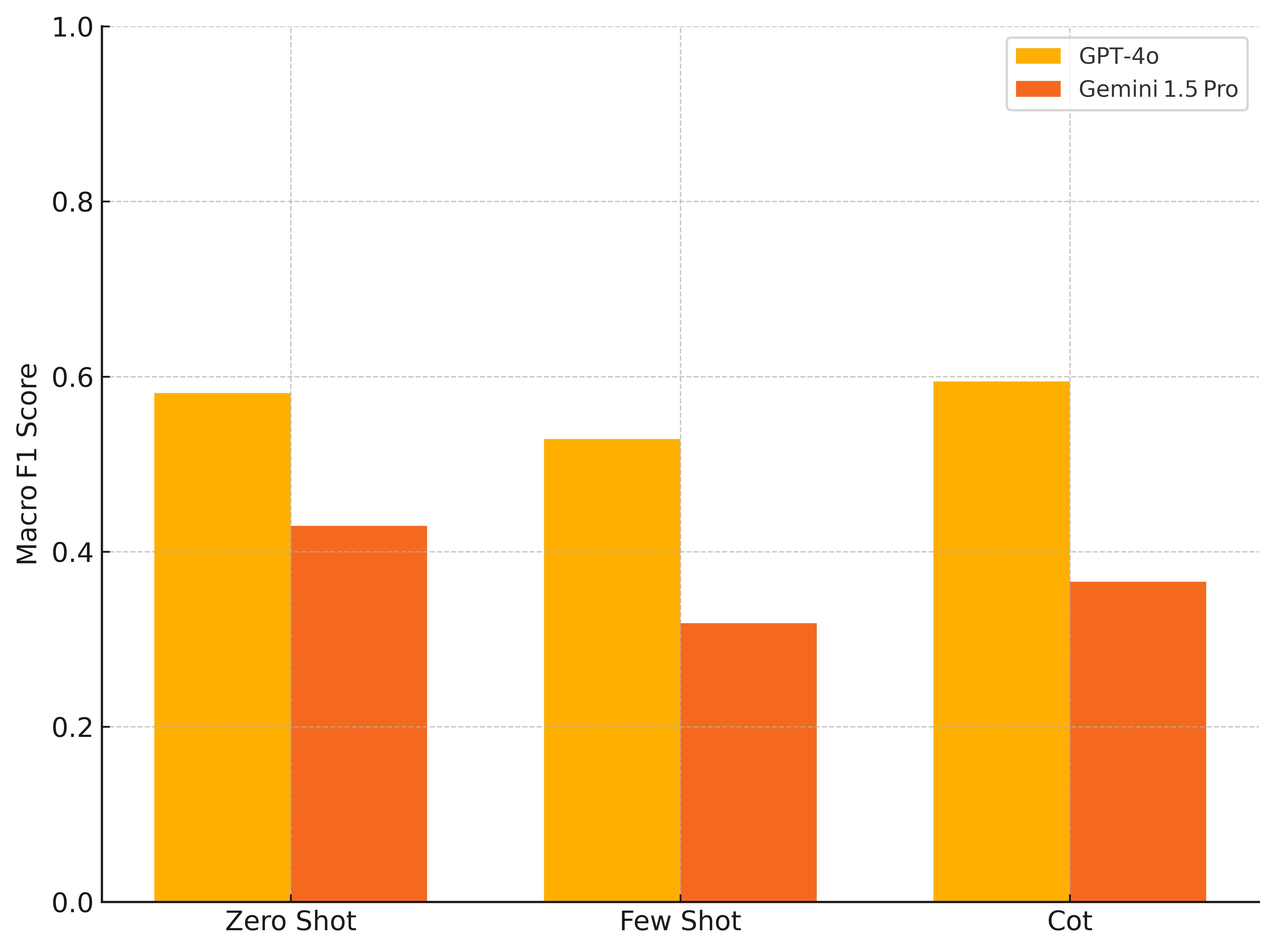}
    \centering
    \caption{Overall emotion recognition performance measured by macro F1-scores across models and prompting strategies.}
    \label{fig:macro_bar}
\end{figure}

Macro F1-scores were used to evaluate overall model performance while accounting for class imbalance across the nine emotional categories, as shown in Figure~\ref{fig:macro_bar}. GPT-4o consistently outperformed Gemini 1.5 Pro across all prompting strategies. GPT-4o achieved macro F1-scores of 57\% (zero-shot), 52\% (few-shot), and 59\% (CoT), while Gemini 1.5 Pro produced macro F1-scores of 43\%, 32\%, and 37\% for the corresponding strategies. Chain-of-thought prompting produced the highest performance for GPT-4o (59\%), representing a 2 percentage point improvement over zero-shot. For Gemini 1.5 Pro, zero-shot prompting achieved the highest macro F1-score (43\%), with few-shot producing the lowest performance across both model architectures.

\subsubsection{Performance by Emotional Category}
\begin{figure}[t]
    \centering
    \scriptsize
    \includegraphics[width=0.60\textwidth]{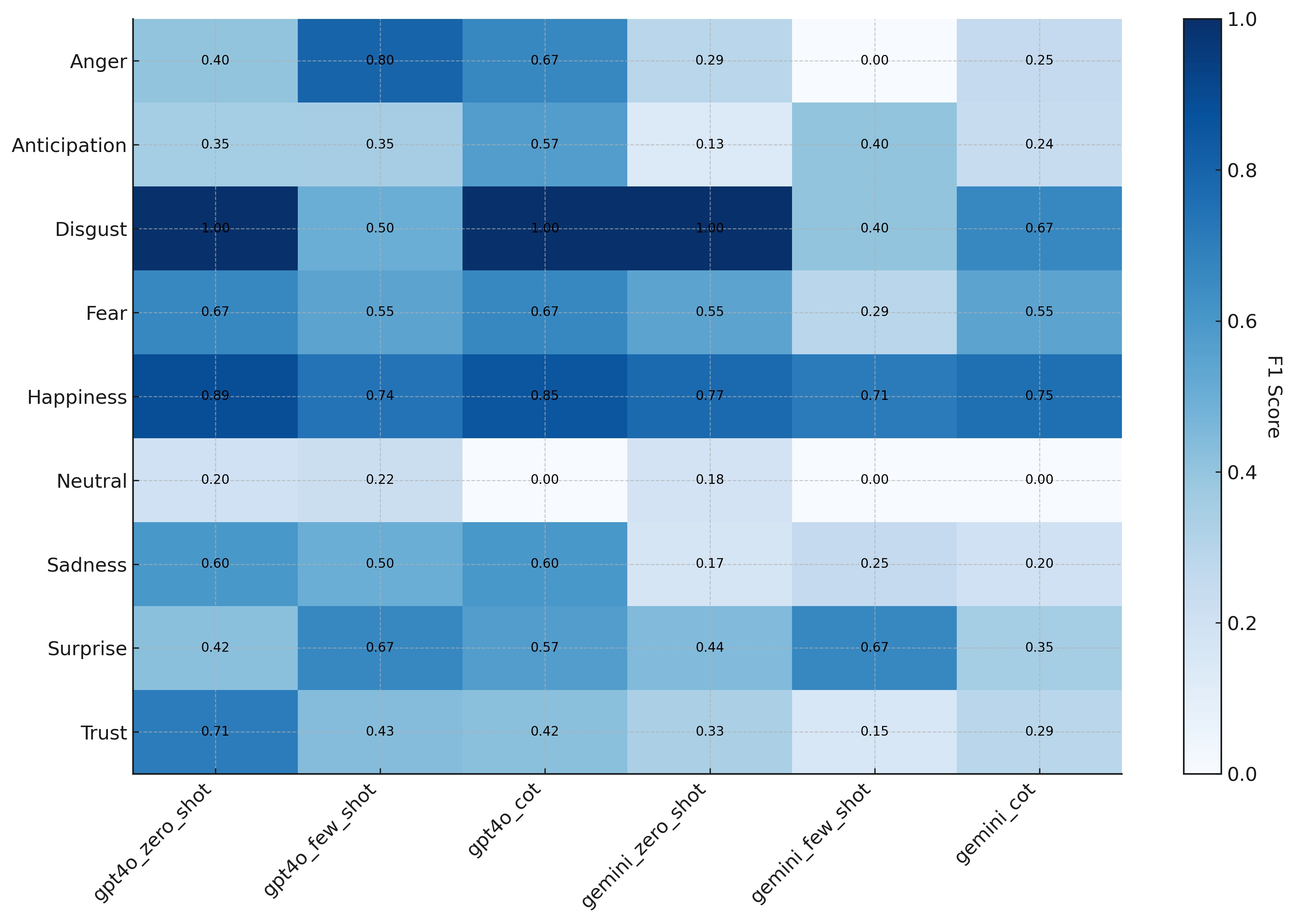}
    \centering
    \caption{Per-emotion F1 scores for GPT-4o and Gemini models across prompting strategies.}
    \label{fig:peremotion_f1}
\end{figure}

Performance varied substantially across emotional categories, with clear distinctions between high-performing and challenging emotions. As shown in Figure~\ref{fig:peremotion_f1}, happiness achieved consistently high F1-scores across all model-prompting combinations, ranging from 69\% (GPT-4o zero-shot) to 85\% (GPT-4o CoT). Trust demonstrated the highest individual performance with GPT-4o zero-shot achieving 71\%, though performance dropped significantly for other configurations. Several emotions proved particularly challenging for both models. Neutral emotion showed consistently poor performance across all configurations, with multiple zero scores for Gemini models and GPT-4o CoT. Anticipation and anger also demonstrated substantial performance gaps between models, with Gemini frequently producing F1-scores below 30\% while GPT-4o maintained moderate performance. Notable performance variability emerged within specific emotions across prompting strategies. For instance, surprise ranged from 35\% (Gemini CoT) to 67\% (GPT-4o few-shot and Gemini few-shot), while disgust showed extreme inconsistency with some configurations achieving 67\% and others dropping to 0\%. Fear and sadness demonstrated more stable performance patterns, with GPT-4o consistently outperforming Gemini across all prompting approaches.

The data reveals that model architecture had a stronger influence on emotion recognition performance than prompting strategy, with GPT-4o showing greater stability and fewer complete classification failures compared to Gemini 1.5 Pro.

\subsection{Systematic Analysis of Emotion Recognition Errors}
Building on the prompting strategy analysis, we conducted a comprehensive examination of misclassification patterns to understand the underlying causes of model errors. We categorized misclassifications into three theoretically-grounded dimensions based on the circumplex model of emotion \citep{russell_circumplex_1980,posner_circumplex_2005}:
\begin{itemize}
\item Valence Inversions: Errors involving confusion between emotions of opposite polarity. The valence dimension represents the positive or negative nature of emotional states. For example, models may confuse positive emotions like happiness ({\scriptsize\<سعادة>}) and trust ({\scriptsize\<ثقة>}) with negative emotions such as sadness ({\scriptsize\<حزن>}) and anger ({\scriptsize\<غضب>}).
\item Arousal Mismatches: Errors where models correctly identify emotional valence but misclassify arousal intensity. High-arousal emotions such as fear ({\scriptsize\<خوف>}), anger ({\scriptsize\<غضب>}), joy ({\scriptsize\<سعادة>}), and surprise ({\scriptsize\<مفاجأة>}) are confused with low-arousal states including sadness ({\scriptsize\<حزن>}), trust ({\scriptsize\<ثقة>}), and neutral expressions ({\scriptsize\<محايد>}) \citep{russell_core_1999, posner_neurophysiological_2008}.
\item Contextual/Cultural Misinterpretations: Cases where models correctly identify both valence and arousal but fail to capture culturally-specific emotional expressions or contextual nuances, resulting in misclassification despite partial dimensional accuracy \citep{jack_facial_2012, barrett_emotional_2019}.
\end{itemize}

\begin{figure}[t]
    \centering
    \scriptsize
    \includegraphics[width=0.43\textwidth]{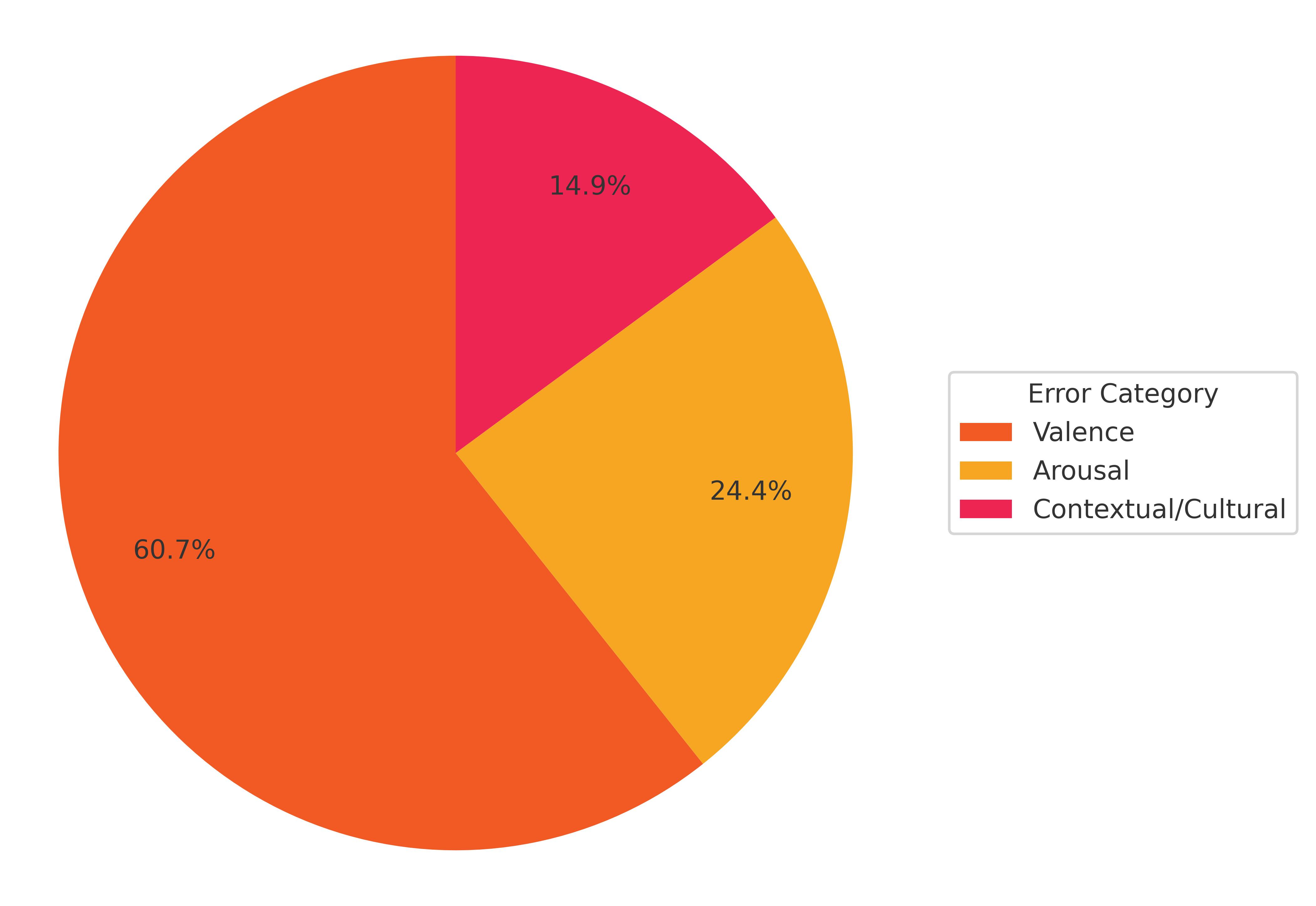}
    \centering
    \caption{Distribution of emotion recognition error types across valence, arousal, and contextual/cultural dimensions.}
    \label{fig:error_cat}
\end{figure}

The error distribution, shown in Figure~\ref{fig:error_cat}, reveals clear patterns in model failures. Valence inversions dominated at 60.7\% (122 out of 201), followed by arousal mismatches at 24.4\% (49 cases) and contextual/cultural misinterpretations at 14.9\% (30 cases). This pronounced imbalance demonstrates systematic rather than random error patterns, with models consistently struggling most with emotional polarity distinctions.

\begin{figure}[t]
    \centering
    \scriptsize
    \includegraphics[width=0.65\textwidth]{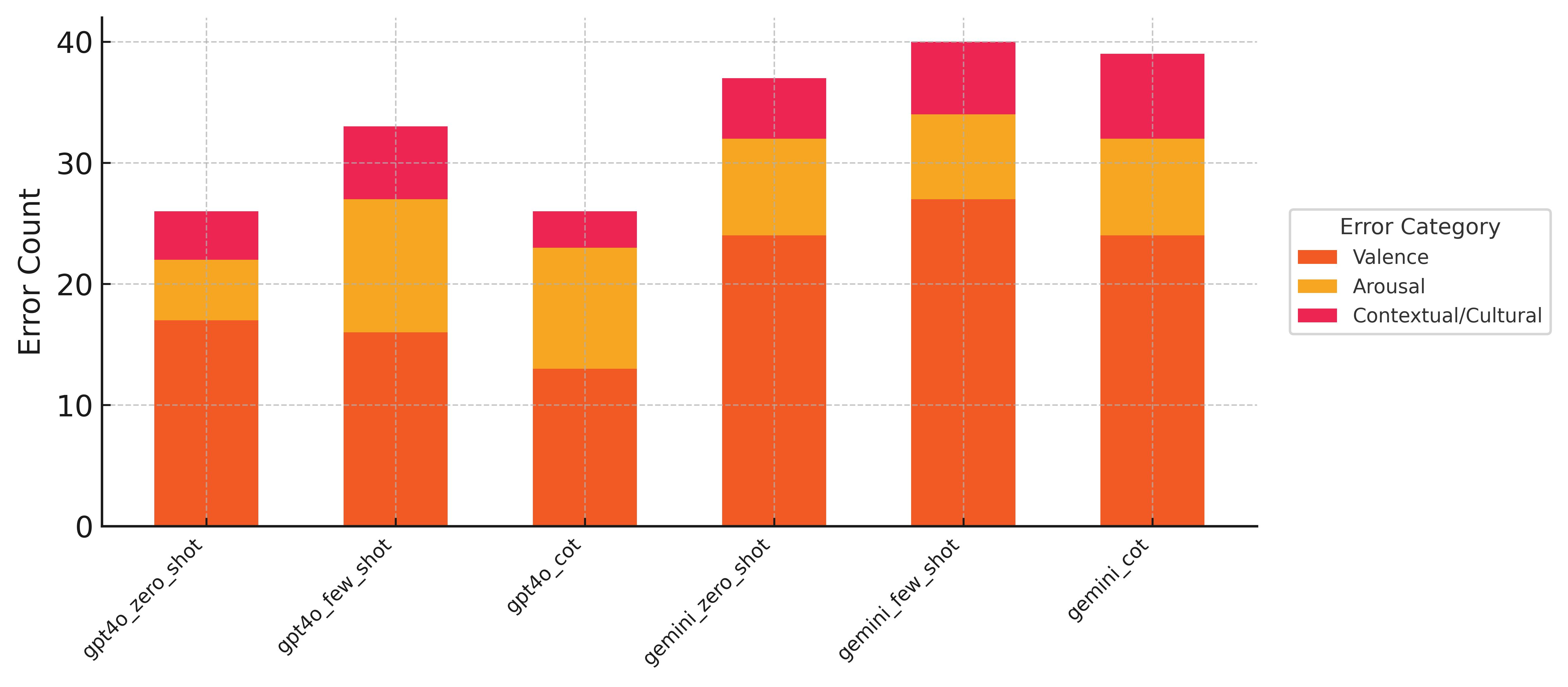}
    \centering
    \caption{Distribution of error types across model-prompting configurations. Error counts are categorized as valence, arousal, or contextual/cultural misclassifications.}
    \label{fig:stacked_error}
\end{figure}

Error distribution patterns varied substantially across model-prompting combinations (Figure~\ref{fig:stacked_error}). GPT-4o demonstrated the lowest total error counts with zero-shot (25 errors) and CoT (26 errors) configurations, while Gemini few-shot produced the highest error count (40 errors).
Valence errors dominated across all configurations, ranging from 48.1\% (GPT-4o CoT) to 70.0\% (Gemini few-shot). Arousal errors showed greater variability, from 20.0\% (GPT-4o zero-shot) to 37.0\% (GPT-4o CoT). Contextual/cultural errors remained relatively consistent across conditions, ranging from 10.0\% to 20.0\% of total errors.
GPT-4o configurations showed more balanced error distributions compared to Gemini, which exhibited higher concentrations of valence-related misclassifications across all prompting strategies.

\begin{figure}[t]
    \centering
    \scriptsize
    \includegraphics[width=0.65\textwidth]{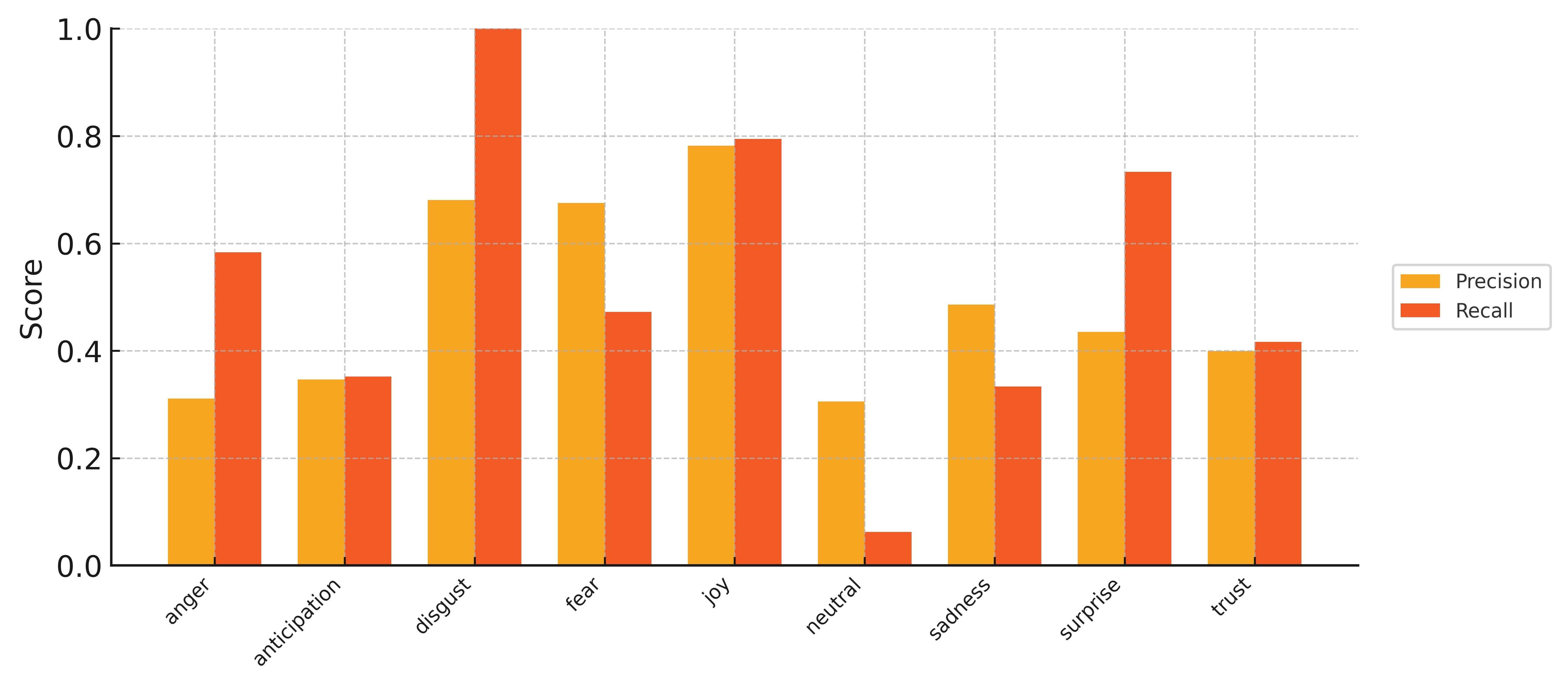}
    \centering
    \caption{Per-emotion precision and recall scores averaged across all model-prompting combinations.}
    \label{fig:precision_recall}
\end{figure}

Precision and recall varied substantially across emotional categories (Figure~\ref{fig:precision_recall}). Joy achieved the highest performance with precision of 78\% and recall of 80\%. Disgust showed high recall (100\%) but moderate precision (68\%). Lower-performing emotions included anger (precision = 32\%, recall = 58\%), anticipation (precision = 35\%, recall = 35\%), and trust (precision = 40\%, recall = 42\%). Neutral emotion demonstrated the poorest performance with precision of 30\% and recall of 6\%.

\subsection{Qualitative Case Analysis}

\begin{figure}[t]
    \centering
    \scriptsize
    \subfloat{\includegraphics[width=5.0cm]{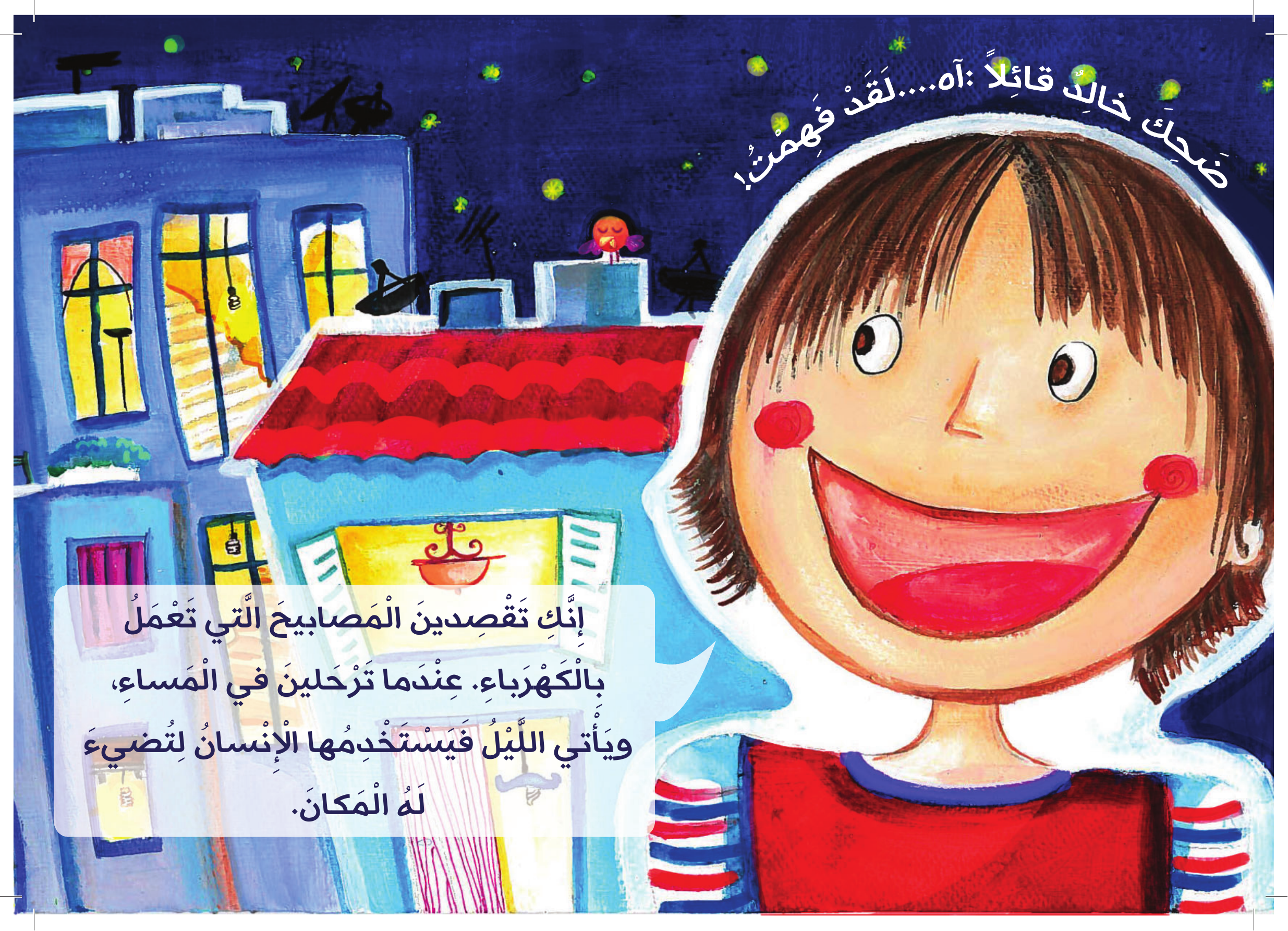}}
    \subfloat{\includegraphics[width=5.0cm]{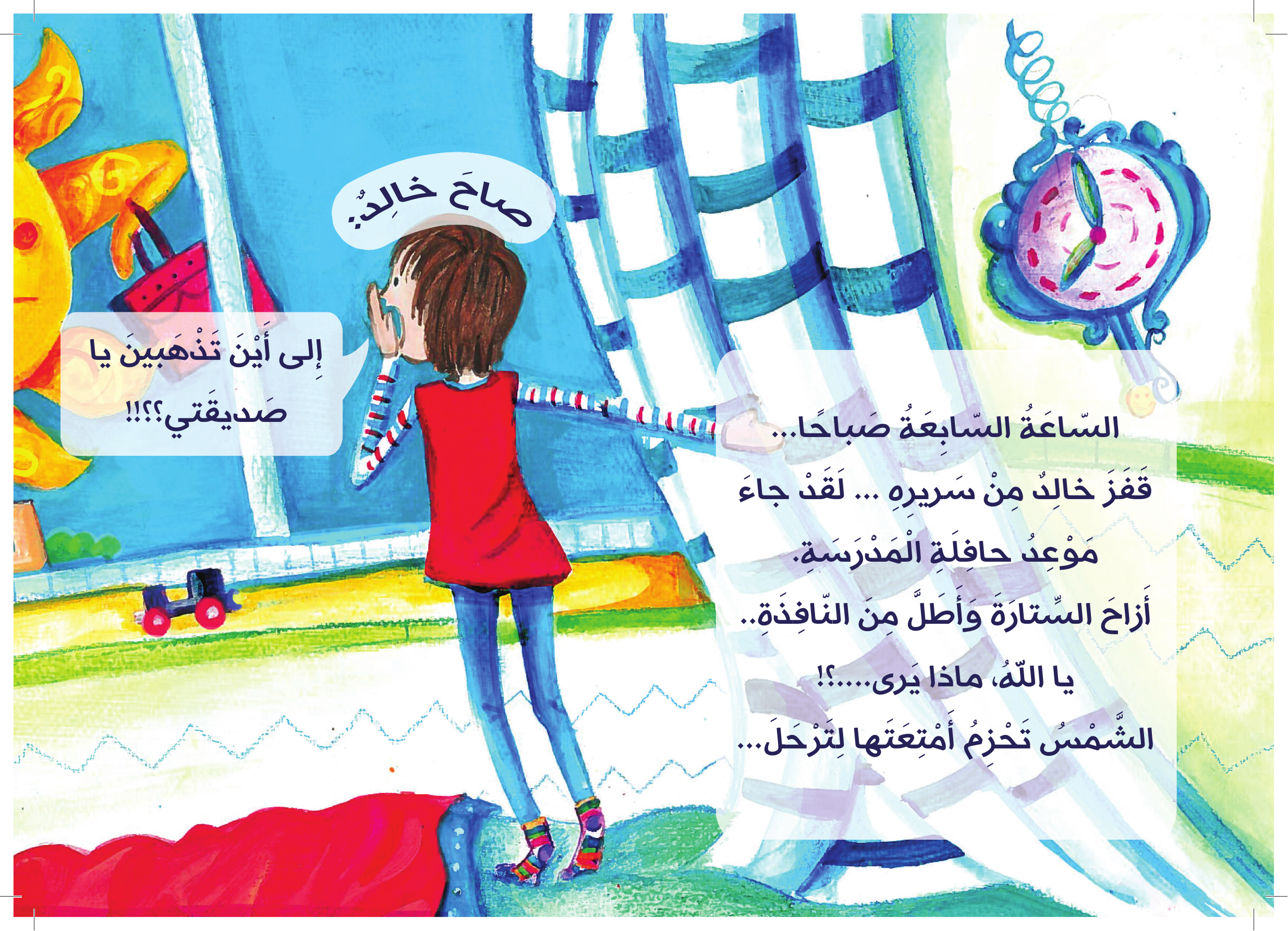}}\\
    \subfloat{\includegraphics[width=5.0cm]{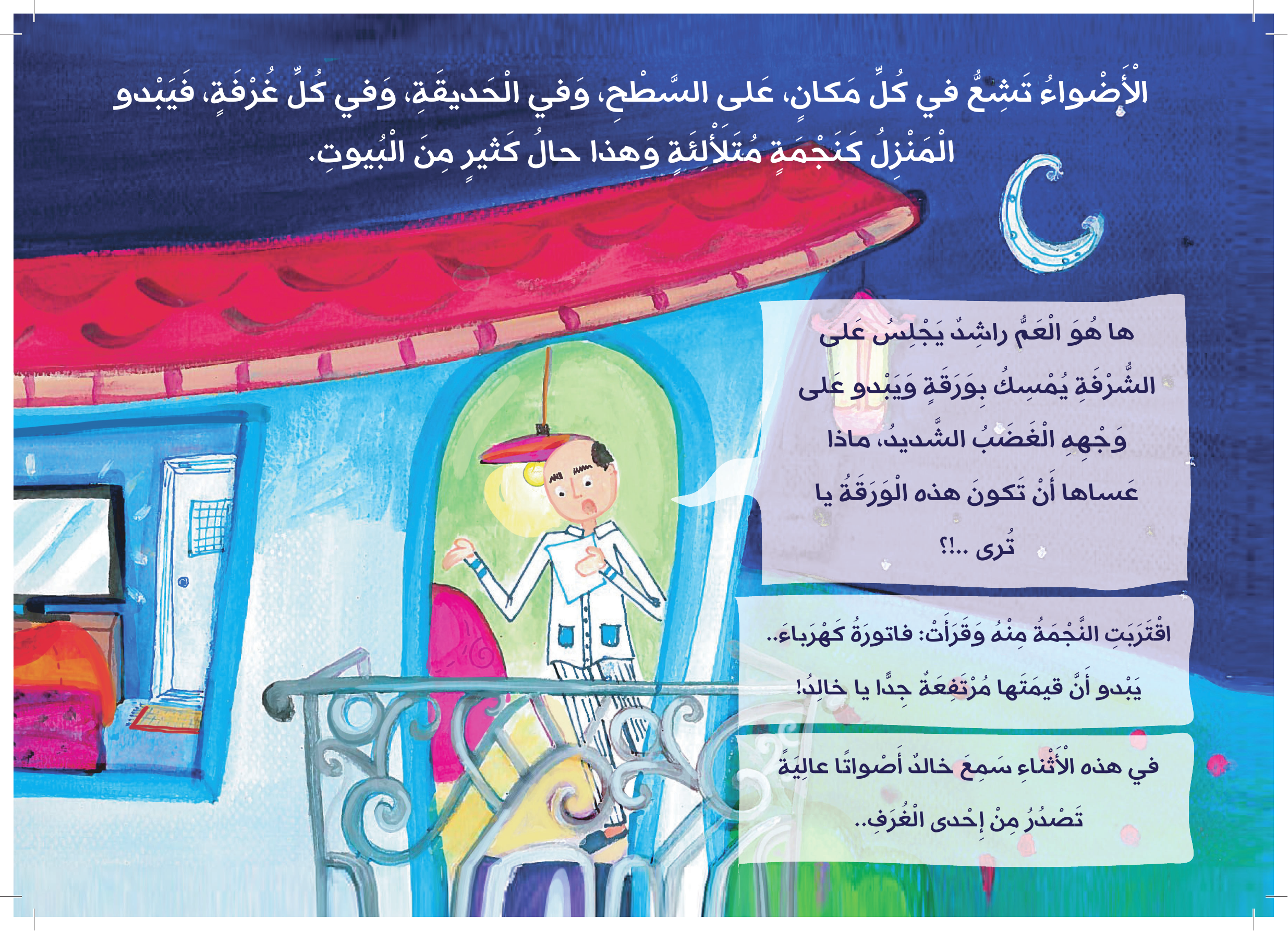}}
    \subfloat{\includegraphics[width=5.0cm]{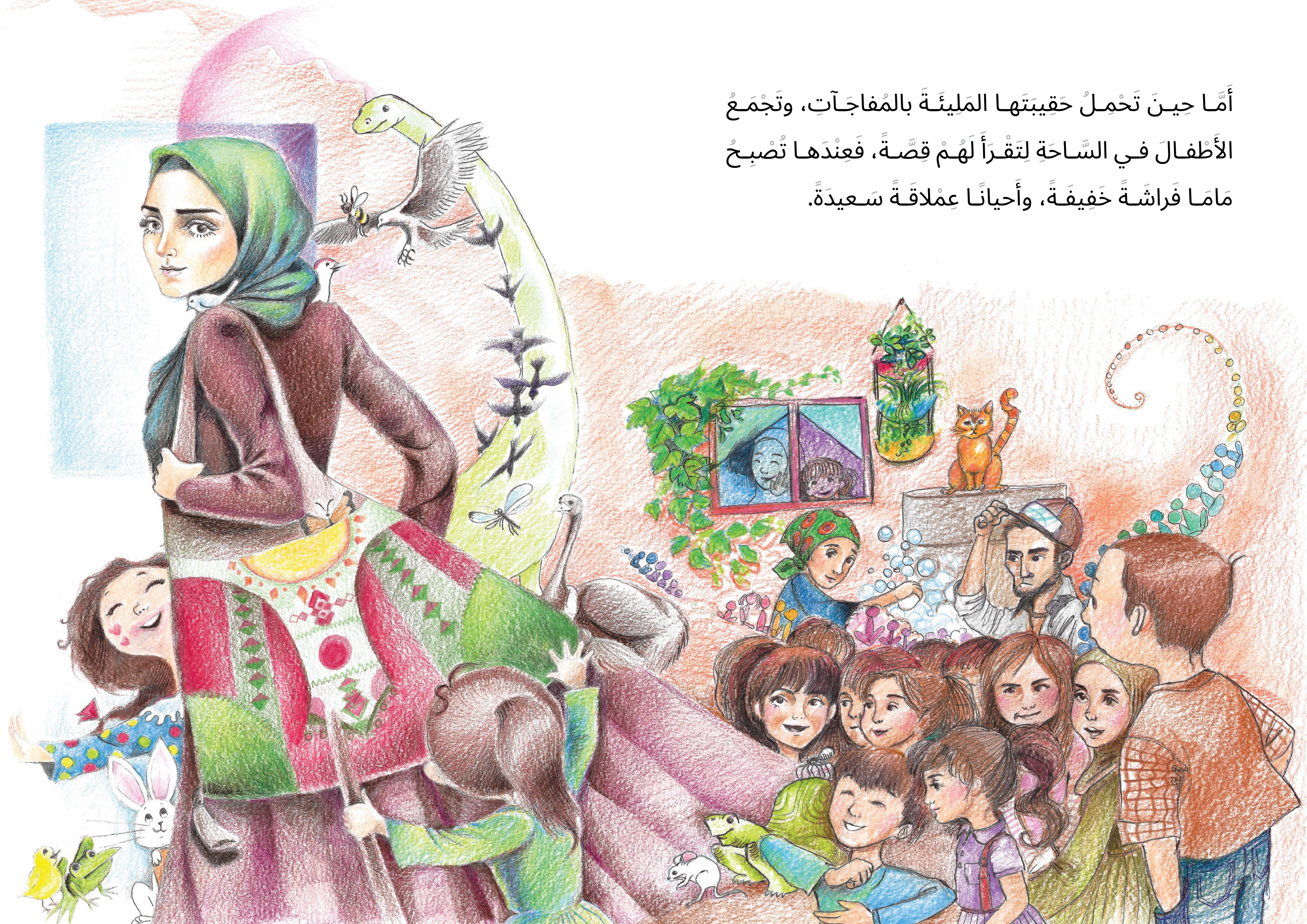}}
    \centering
    \caption{Qualitative case examples showing different model-human agreement patterns. (\textbf{a}) High agreement scenario where all models correctly identified clear emotional indicators (happiness). (\textbf{b}) Ambiguous visual cues leading to divergent model classifications despite human consensus. (\textbf{c}) Text-visual interaction case where most models predicted anger while humans annotated the emotion as neutral. (\textbf{d}) Systematic valence inversion case where GPT-4o agreed with human happiness annotation while all Gemini configurations predicted negative emotions. Cases demonstrate varying challenges in Arabic emotion recognition across different visual and contextual scenarios.}
    \label{fig:case_panels}
\end{figure}

To complement our quantitative analysis, we examined some representative image cases to understand how visual clarity and contextual factors influence model performance (Figure~\ref{fig:case_panels}).

\paragraph{Case 1: High Agreement on Clear Emotional Cues}
Panel (a) shows an image where the human annotation was happiness ({\scriptsize\<سعادة>}), and all models across all prompting strategies correctly identified this emotion. The image features a child with a prominent smile and bright, cheerful visual elements. This case demonstrates successful emotion recognition when clear visual indicators are present.

\paragraph{Case 2: Ambiguous Visual Cues Leading to Divergent Classifications}
Panel (b) was human-annotated as neutral ({\scriptsize\<محايد>}), but models disagreed in their classifications. GPT-4o predicted surprise ({\scriptsize\<مفاجأة>}), while Gemini few-shot predicted fear ({\scriptsize\<خوف>}). The image shows more subtle facial expressions and less distinct emotional markers compared to Panel (a), resulting in varied model interpretations.

\paragraph{Case 3: Text and Context Override Visual Cues}
Panel (c) was human-labeled as neutral ({\scriptsize\<محايد>}), but five out of six model configurations predicted anger ({\scriptsize\<غضب>}). The Arabic text content appears to describe conflict and an anger emotion, which might have influenced model predictions despite the character's calm visual appearance.

\paragraph{Case 4: Systematic Valence Inversion}
Panel (d) demonstrates architectural differences in valence processing. Human annotation and all GPT-4o configurations identified happiness ({\scriptsize\<سعادة>}), while all Gemini configurations systematically predicted negative emotions: sadness and fear ({\scriptsize\<حزن>}, {\scriptsize\<خوف>}).

\subsection{Human-AI Alignment}
To assess alignment between large language models and human annotators in emotion recognition from children's storybooks, we analyzed agreement rates across six model-prompting combinations compared against human-labeled ground truth using Cohen's Kappa statistics.

Results revealed substantial differences in human-AI alignment across models and prompting techniques (Table~\ref{tab:cohens_kappa}). GPT-4o consistently demonstrated higher alignment with human annotations, achieving Cohen's Kappa values of 0.56 (zero-shot), 0.46 (few-shot), and 0.56 (CoT). These values indicate moderate agreement according to standard interpretation guidelines. Gemini showed notably lower performance with Cohen's Kappa values of 0.37 (zero-shot), 0.31 (few-shot), and 0.34 (CoT), indicating fair agreement levels. Prompting strategies produced different effects across models. For GPT-4o, zero-shot and CoT achieved equally high performance ($\kappa = 0.56$), while few-shot showed reduced alignment ($\kappa = 0.46$). For Gemini, zero-shot prompting yielded the highest agreement ($\kappa = 0.37$), followed by CoT ($\kappa = 0.34$) and few-shot ($\kappa = 0.31$). These patterns suggest that elaborate prompting strategies do not uniformly enhance human-AI alignment across different model architectures.

The consistent performance gap between GPT-4o and Gemini across all prompting strategies (0.15--0.22 $\kappa$ difference) indicates systematic differences in human-AI alignment capabilities. GPT-4o maintained moderate agreement levels across all conditions, while Gemini consistently achieved only fair agreement, suggesting fundamental differences in emotion recognition approaches between the two architectures.

\begin{table}[H]
\caption{Cohen's Kappa values for human-AI agreement across model-prompting combinations.}
\label{tab:cohens_kappa}
\centering
\begin{tabular}{lccc}
\toprule
\textbf{Model} & \textbf{Zero-shot} & \textbf{Few-shot} & \textbf{CoT} \\ 
\midrule
\textbf{GPT-4o} & 0.56 & 0.46 & 0.56 \\ 
\midrule
\textbf{Gemini} & 0.37 & 0.31 & 0.34 \\ 
\bottomrule
\end{tabular}
\end{table}

\begin{figure}[t]
    \centering
    \scriptsize
    \includegraphics[width=0.50\textwidth]{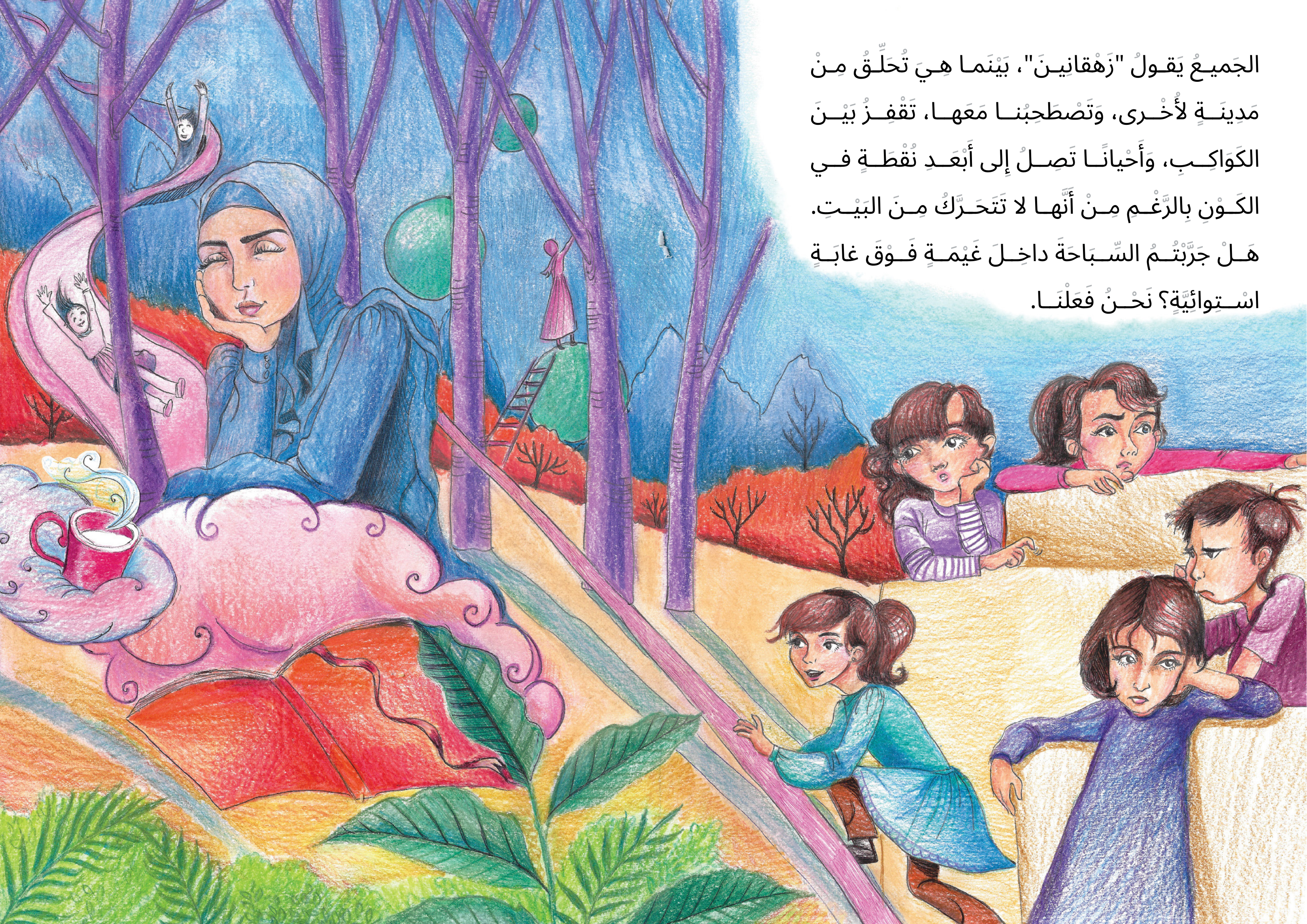}
    \centering
    \caption{Multi-character narrative scene illustrating the complexity of emotion recognition when multiple facial expressions and cultural storytelling elements are present. Both images required integration of visual and narrative context that proved challenging for current multimodal language models.}
    \label{fig:mama_panel}
\end{figure}

We identified a subset of images where all six model-prompting combinations diverged from human-labeled emotions. These cases demonstrated high interpretive complexity, often involving multiple characters with distinct emotional expressions, embedded symbolic elements, or contextual dependencies tied to previous narrative pages. Examples include images such as Figure~\ref{fig:mama_panel} that consistently revealed disagreement across all configurations, indicating intrinsic ambiguity in emotional interpretation. Analysis of these discrepancies showed that models were more prone to misclassification when emotion recognition required integrating both textual context and visual semantics. This pattern underscores the challenges faced by current models when processing illustrated multimodal content that requires narrative continuity.

\begin{figure}[t]
    \centering
    \scriptsize
    \includegraphics[width=0.60\textwidth]{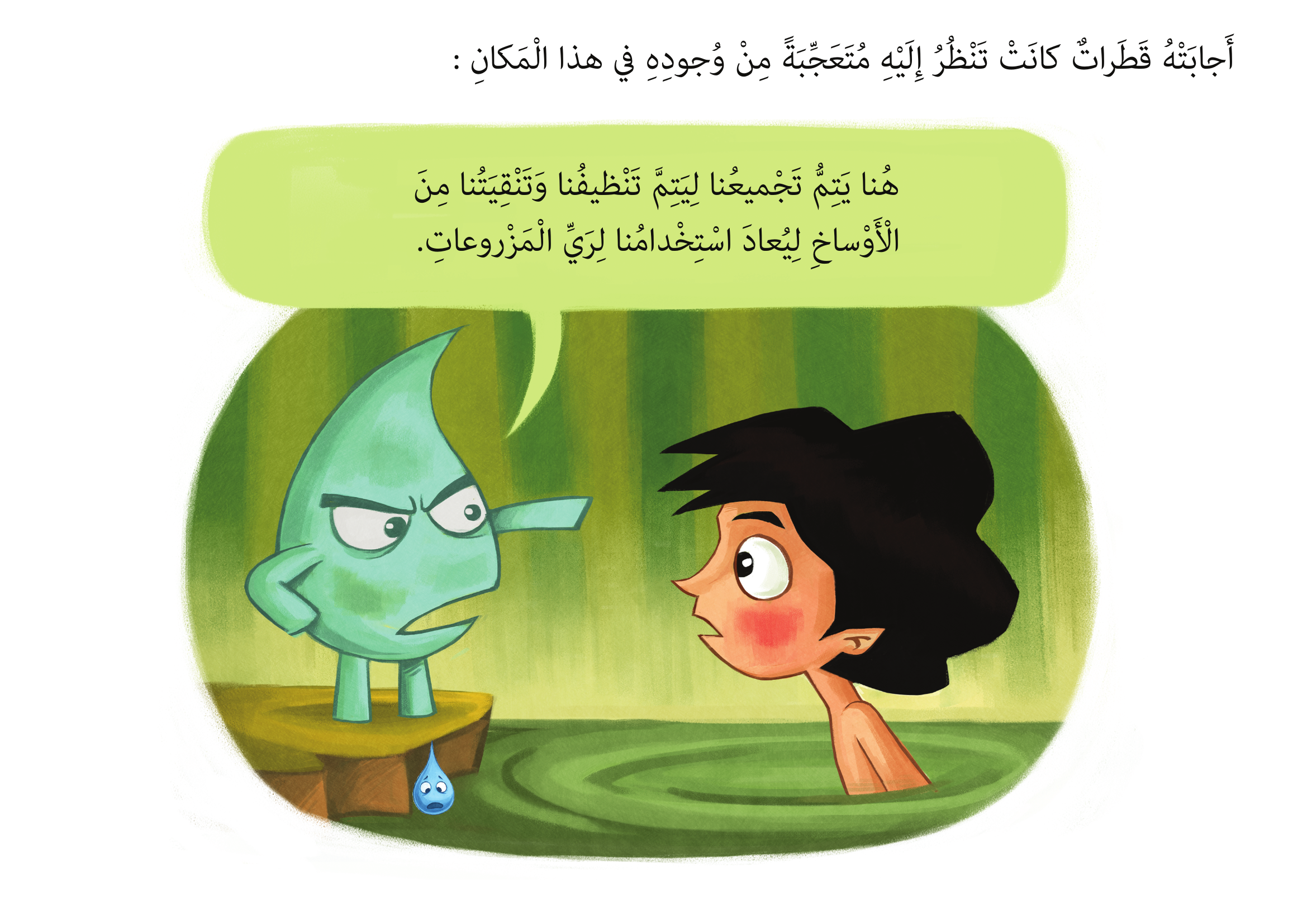}
    \centering
    \caption{Character dialogue scene demonstrating challenges in interpreting emotional context between personified and human characters.}
    \label{fig:open_tab}
\end{figure}

Agreement rates varied substantially across emotional categories. Certain emotions, such as happiness ({\scriptsize\<سعادة>}) and surprise ({\scriptsize\<مفاجأة>}), achieved more consistent recognition across model-prompting combinations compared to others like anticipation ({\scriptsize\<ترقب>}) and neutral ({\scriptsize\<محايد>}). COT prompting showed inconsistent effects on human alignment, sometimes improving agreement in specific cases while reducing it in others (Figure~\ref{fig:open_tab}), indicating that prompting effects vary substantially based on image content and narrative context.

\subsection{Supplementary Analysis Results}
To investigate factors influencing emotion recognition performance, we conducted two supplementary experiments on a representative subset (n=8, 10.7\%) of our dataset. For baseline comparison, these same 8 images achieved 37.5\% correct classification (3/8) for GPT-4o and 12.5\% (1/8) for Gemini across all prompting strategies in the main experiment. In the character-focused segmentation experiment, we isolated individual characters from their narrative contexts. This produced divergent effects across models: GPT-4o's performance dropped to 12.5\% (1/8) across all prompting strategies—a 25 percentage point decrease—while exhibiting a strong bias toward "anticipation" predictions (37.5\% of all responses). Conversely, Gemini's performance improved to 25\% (2/8) for zero-shot and few-shot—a 12.5 percentage point increase—though CoT remained at baseline (12.5\%). Despite this improvement, Gemini defaulted to high-arousal emotions, particularly "anger" (33.3\%) and "surprise" (29.2\%). Notably, positive emotions showed complete recognition failure: both happiness instances and the single trust instance were misclassified by all model configurations, while surprise was correctly identified in 5 out of 6 attempts across all models.

The Plutchik's wheel augmentation experiment produced similarly mixed results. GPT-4o's performance decreased to 25\% (2/8) across all prompting strategies—a 12.5 percentage point decline from baseline. Gemini showed varied responses: zero-shot maintained baseline performance (12.5\%), few-shot dropped to complete failure (0/8), while CoT improved to 25\% (2/8)—doubling its baseline performance. Rather than improving classification accuracy, the expanded emotional vocabulary led to overcomplicated predictions, with models introducing 13 unique emotion labels instead of the original 9. GPT-4o exhibited a pattern of over-sophistication, labeling basic sadness as "contemplation" ({\scriptsize\<تأمل>}) and neutral states as "love" ({\scriptsize\<حب>}), while Gemini showed intensity escalation, replacing surprise with "astonishment" ({\scriptsize\<دهشة>}) and anger with "contempt" ({\scriptsize\<ازدراء>}). These supplementary findings reveal model-specific sensitivities: GPT-4o appears to rely heavily on holistic scene context, while Gemini can benefit from focused attention or explicit taxonomic scaffolding, though at the cost of emotional granularity appropriate for children's literature.


\section{Discussion}

\subsection{Architectural Differences in Multimodal Emotion Processing}
The consistent performance advantage of GPT-4o over Gemini across all conditions suggests fundamental differences in how these architectures integrate visual and linguistic emotional information. This gap likely reflects variations in training methodologies, model scale, and multimodal fusion strategies \citep{li_blip_2022}. GPT-4o's superior stability across prompting strategies indicates more robust internal representations of emotional concepts, potentially due to more sophisticated attention mechanisms or better-calibrated visual encoders \citep{team_gemini_2023}. 

The text-visual interaction effects observed in our qualitative analysis further illustrate these architectural differences. Case 3 revealed that models frequently prioritize Arabic textual content over visual emotional cues, with five out of six configurations predicting anger despite neutral facial expressions when conflict-related text was present. This suggests varying capabilities in balancing multimodal information sources across different model architectures.

The failure of few-shot prompting to improve performance for either model challenges conventional assumptions about in-context learning for emotion recognition. This suggests that emotion classification may require different cognitive processes than typical few-shot tasks, possibly because emotional interpretation depends more on learned associations than pattern matching from examples \citep{dong_survey_2024}.

\subsection{The Valence Processing Deficit}
The overwhelming prevalence of valence errors (60.7\%) reveals a critical limitation in current MLLMs' understanding of emotional polarity. This finding suggests that models may process emotions as discrete categories rather than understanding the underlying dimensional structure of affect \citep{russell_circumplex_1980, posner_circumplex_2005}. The dominance of valence over arousal errors indicates that models struggle more with the fundamental positive-negative distinction than with intensity judgments. 

This pattern aligns with psychological theories suggesting that valence processing requires deeper semantic understanding and cultural knowledge than arousal detection \citep{barrett_conceptual_2014}. The models' difficulty with valence may reflect their reliance on surface-level visual features rather than contextual understanding of emotional meaning within cultural frameworks.

Our qualitative analysis reinforces these theoretical insights. Case 4 demonstrates how architectural differences manifest in systematic valence inversion, where GPT-4o correctly identified happiness while all Gemini configurations predicted negative emotions (sadness, fear) for the same image. This pattern exemplifies how valence processing deficits operate consistently within model architectures rather than occurring randomly.

\subsection{Cultural and Contextual Challenges}
The models' struggle with culturally embedded emotions highlights the limitations of predominantly Western-trained AI systems when applied to Arabic contexts \citep{bender_dangers_2021, blodgett_language_2020}. The systematic nature of misclassifications suggests that current training paradigms inadequately capture culture-specific emotional expressions and social contexts that influence affective interpretation \citep{ elfenbein_cultural_2003}.

The poor performance on neutral emotions reveals a particular challenge for AI systems: distinguishing between the absence of clear emotional signals and the presence of genuinely neutral states \citep{barrett_emotional_2019}. This difficulty may stem from models' tendency to over-interpret visual information, seeking emotional content even in ambiguous scenarios.

\subsection{Prompting Strategy Implications}
The mixed effects of chain-of-thought prompting suggest that elaborate reasoning may not uniformly benefit emotion recognition tasks. For GPT-4o, CoT sometimes led to over-interpretation of narrative context, while for Gemini, it often increased inconsistency. This indicates that emotion recognition may benefit from more intuitive, system-1 type processing rather than deliberative reasoning, reflecting how humans often process emotional information rapidly and automatically. The prompting effects also suggest that different model architectures may require tailored interaction strategies, with standardized prompting approaches potentially suboptimal across diverse AI systems.

\subsection{Theoretical Implications for Affective AI}
Our findings challenge the assumption that larger, more sophisticated language models automatically excel at emotion recognition \citep{picard_affective_1997}. The systematic error patterns suggest that current training approaches may not adequately develop the multimodal integration and cultural understanding necessary for robust emotional AI systems.

The dominance of valence errors indicates that developing AI systems with better emotional intelligence requires moving beyond surface-level pattern recognition toward deeper understanding of affective meaning and cultural context \citep{mohammad_understanding_2018}. This suggests a need for training paradigms that explicitly model emotional dimensions rather than treating emotions as discrete, isolated categories \citep {abdul-mageed_emonet_2017}.

For Arabic literacy applications, these findings have direct implications for educational technology deployment. Current MLLMs require careful prompt engineering and potentially specialized fine-tuning before implementation in Arabic educational contexts. The systematic nature of valence errors suggests that emotion-aware educational systems should incorporate bias detection and correction mechanisms, particularly when processing culturally-specific emotional content.

\subsection{Implications of Contextual and Taxonomic Constraints}
Our supplementary analyses reveal fundamental differences in how current MLLMs process emotion in narrative contexts. The character segmentation experiment produced strikingly divergent results: GPT-4o's performance dropped from 37.5\% to 12.5\% (a 25 percentage point decrease), while Gemini's performance improved from 12.5\% to 25\% for zero-shot and few-shot approaches. This bidirectional effect suggests contrasting architectural dependencies—GPT-4o appears to rely heavily on holistic scene processing, integrating background elements and interpersonal dynamics into its emotion recognition, while Gemini may suffer from visual complexity and benefit from focused attention on facial features. The complete failure to recognize positive emotions (0\% for happiness and trust) across both models, versus preserved surprise recognition (83\%), reinforces that culturally-expressed emotions like happiness depend on scenic elements (colors, spatial relationships, shared activities) rather than facial features alone.

The Plutchik's wheel experiment similarly revealed model-specific responses to theoretical scaffolding. While GPT-4o's performance declined from 37.5\% to 25\%, Gemini's CoT actually improved from 12.5\% to 25\%, though few-shot catastrophically dropped to 0\%. This suggests that explicit taxonomic frameworks can stabilize weaker baselines but may interfere with stronger models' learned representations. The models' introduction of sophisticated emotions like "contemplation" for basic sadness or "love" for neutral states reveals what we term "theoretical interference"—where abstract psychological frameworks override practical pattern recognition. Critically, both experiments demonstrate that no single approach optimizes performance across models: GPT-4o requires complete scenes without theoretical scaffolding, while Gemini can benefit from constrained focus or explicit ontological cues, though at the cost of nuanced interpretation.

These findings challenge universal approaches to emotion recognition in educational AI. Rather than seeking optimal preprocessing or prompting strategies, our results suggest the need for model-adaptive pipelines that leverage each architecture's strengths. For GPT-4o, this means preserving full narrative context; for Gemini, selective attention or taxonomic guidance may improve performance on specific tasks. However, the persistent failure on positive emotions and the inappropriate sophistication introduced by Plutchik's framework underscore that emotion in children's literature serves pedagogical rather than psychological functions. Arabic educational AI development must therefore prioritize culturally-grounded, context-preserving approaches that recognize emotions as narrative devices rather than isolated psychological states.

\subsection{Limitations}
Several methodological constraints should be acknowledged in interpreting our findings. First, we evaluated models at a specific point in time, and the rapid evolution of GPT-4o and Gemini 1.5 means that model capabilities may change with subsequent updates, potentially affecting the long-term relevance of our results. Second, these models exhibit inherent variability and sensitivity to prompt variations that could influence results, even when using consistent formats. Finally, our study lacked comparison with specialized Arabic NLP systems or purpose-built emotion recognition models, which might provide valuable performance baselines beyond general-purpose models. Future evaluations would benefit from including comparisons with specialized Arabic NLP tools \citep{abdul-mageed_emonet_2017} or emotion recognition systems trained specifically for affective computing tasks.

\subsubsection{Dataset Constraints}
We analyzed 75 images systematically extracted from seven Arabic storybooks, with a minimum of ten illustrations from each source. A significant limitation was the highly imbalanced distribution of emotional labels in our dataset. Happiness ({\scriptsize\<سعادة>}) comprised 40\% of all annotations (30 out of 75 images), while emotions such as anger ({\scriptsize\<غضب>}) and disgust ({\scriptsize\<قرف>}) were severely underrepresented at 2.7\% (2 images) and 1.3\% (1 image) respectively. Although we used macro F1-scores to account for class imbalance in evaluation metrics, the limited examples of rare emotions reduce statistical reliability and generalizability of findings for these categories.

The restricted sample size and imbalanced class distribution limit the statistical power and generalizability of our findings, particularly for underrepresented emotional categories. Furthermore, our emotion annotation framework relied on Plutchik's taxonomy supplemented with a neutral classification, which may not fully capture the complexity or culturally-specific aspects of emotions depicted in Arabic visual narratives \citep{plutchik_nature_2001}. Although we employed four native Arabic-speaking annotators and implemented a structured consensus procedure, subjective interpretations in emotionally ambiguous cases cannot be entirely eliminated \citep{hovy_social_2016}.

\subsection{Future Research Directions}
Future work should expand the dataset to include a broader corpus of Arabic storybooks representing diverse visual styles and cultural contexts, with particular emphasis on balancing emotional categories to ensure adequate representation of less common emotions. Incorporating more complex emotional categories beyond Plutchik's framework, such as embarrassment, pride, or culturally-specific emotional concepts, would enhance the validity of the evaluation approach. Applying explainable AI (XAI) techniques \citep{ribeiro_why_2016} could reveal reasoning patterns behind model misclassifications, particularly in ambiguous narrative contexts where chain-of-thought prompting demonstrated inconsistent performance. Developing automated annotation and evaluation systems would improve research scalability in Arabic literacy applications.

From an educational technology perspective, developing culturally-adaptive prompting strategies specifically for Arabic educational content represents a critical research direction. Such strategies should account for the text-visual interaction effects observed in our analysis and provide frameworks for detecting and mitigating systematic valence biases in educational applications.

Additionally, fine-tuning multilingual models on culturally-specific Arabic emotional content could significantly improve their sensitivity to subtle emotional cues and cultural nuances. Finally, educational intervention studies examining how these emotion-aware systems impact actual learning outcomes in Arabic literacy development would provide valuable insights for practical implementation in educational settings.


\section{Conclusion}
This study provides the first systematic evaluation of multimodal large language models for Arabic emotion recognition in children's storybook illustrations. GPT-4o consistently outperformed Gemini 1.5 across all prompting strategies, achieving macro F1-scores of 57-59\% compared to Gemini's 32-43\%. Human-AI alignment showed similar patterns, with GPT-4o maintaining moderate agreement ($\kappa = 0.46-0.56$) versus Gemini's fair agreement ($\kappa = 0.31-0.37$).
Error analysis revealed systematic patterns rather than random failures, with valence inversions dominating at 60.7\% of misclassifications. Both models struggled with culturally nuanced emotions and neutral states, indicating fundamental limitations in processing affective content within Arabic cultural contexts.
For educational applications, we recommend zero-shot or chain-of-thought prompting with GPT-4o, while avoiding few-shot approaches that consistently underperformed. Future work should prioritize culturally responsive training data and enhanced valence processing to develop more effective emotionally intelligent educational technologies for Arabic-speaking learners.

\bibliographystyle{unsrt}  
\bibliography{references}

\end{document}

%% file: tex/abstract.tex
\begin{abstract}
    Emotion recognition capabilities in multimodal AI systems are crucial for developing culturally responsive educational technologies, yet remain underexplored for Arabic language contexts where culturally appropriate learning tools are critically needed. This study evaluates the emotion recognition performance of two advanced multimodal large language models, GPT-4o and Gemini 1.5 Pro, when processing Arabic children's storybook illustrations. We assessed both models across three prompting strategies (zero-shot, few-shot, and chain-of-thought) using 75 images from seven Arabic storybooks, comparing model predictions with human annotations based on Plutchik's emotional framework. GPT-4o consistently outperformed Gemini across all conditions, achieving the highest macro F1-score of 59\% with chain-of-thought prompting compared to Gemini's best performance of 43\%. Error analysis revealed systematic misclassification patterns, with valence inversions accounting for 60.7\% of errors, while both models struggled with culturally nuanced emotions and ambiguous narrative contexts. These findings highlight fundamental limitations in current models' cultural understanding and emphasize the need for culturally sensitive training approaches to develop effective emotion-aware educational technologies for Arabic-speaking learners.
\end{abstract}